\documentclass[final]{cvpr}

\usepackage{times}
\usepackage{epsfig}
\usepackage{graphicx}
\usepackage{amsmath}
\usepackage{amssymb}
\usepackage{adjustbox}
\usepackage{booktabs}
\usepackage{multirow}
\usepackage{bm}
\usepackage[sort&compress,square,comma,numbers]{natbib}
\makeatletter
\renewcommand\bibsection%
{
 \section*{\refname
    \@mkboth{\MakeUppercase{\refname}}{\MakeUppercase{\refname}}}
}
\makeatother

\usepackage[pagebackref=true,breaklinks=true,colorlinks,bookmarks=false]{hyperref}

\def\cvprPaperID{3675} 
\def\confYear{CVPR 2021}

\begin{document}

\title{Metadata Normalization}

\author{Mandy Lu, Qingyu Zhao, Jiequan Zhang, Kilian M. Pohl, Li Fei-Fei, Juan Carlos Niebles, Ehsan Adeli\\
Stanford University, Stanford, CA 94305\\
{\tt\small \{mlu,eadeli\}@cs.stanford.edu}\\
{\small~~\url{https://mml.stanford.edu/MDN/}}
%
}

\maketitle

\begin{abstract}
Batch Normalization (BN) and its variants have delivered tremendous success in combating the covariate shift induced by the training step of deep learning methods. While these techniques normalize feature distributions by standardizing with batch statistics, they do not correct the influence on features from extraneous variables or multiple distributions. Such extra variables, referred to as metadata here, may create bias or confounding effects (\eg, race when classifying gender from face images). We introduce the Metadata Normalization (MDN) layer, a new batch-level operation which can be used end-to-end within the training framework, to correct the influence of metadata on feature distributions. MDN adopts a regression analysis technique traditionally used for preprocessing to remove (regress out) the metadata effects on model features during training. We utilize a metric based on distance correlation to quantify the distribution bias from the metadata and demonstrate that our method successfully removes metadata effects on four diverse settings: one synthetic, one 2D image, one video, and one 3D medical image dataset. 
\end{abstract}

\section{Introduction}

Recent advances in fields such as computer vision, natural language processing, and medical imaging have been propelled by tremendous progress in deep learning \cite{bahri2020statistical}. Deep neural architectures owe their success to their large number of trainable parameters, which encode rich information from the data. However, since the learning process can be extremely unstable, much effort is spent on carefully selecting a model through hyperparameter tuning, an integral part of approaches such as \cite{ioffe2015batch,maclaurin2015gradient}. To aid with model development, normalization techniques such as Batch Normalization (BN) \cite{ioffe2015batch} and Group Normalization (GN) \cite{wu2018group} make the training process more robust and less sensitive to covariate or distribution shift. 

BN and GN perform feature normalization by standardizing them solely using batch or group statistics (\ie, mean and standard deviation). Although they have pushed the state-of-the-art forward, they do not handle extraneous dependencies in the data other than the input and output label variables. In many applications, confounders \cite{zhaoadeli2020cf-net,wang2019removing} or protected variables \cite{salimi2019interventional} (sometimes referred to as bias variables \cite{adeli2019representation}) may inject bias into the learning process and skew the distribution of the learned features. For instance, (i) when training a gender classification model from face images, an individual's race (quantified by skin shade) has a crucial influence on prediction performance as shown in \cite{buolamwini2018gender}; (ii) in video understanding, action recognition models are often driven by the scene \cite{Huang2018WhatMA,choi2019why} instead of learning the harder movement-related action cues; (iii) for medical studies, patient demographic information or data acquisition site location (due to device and scanner differences) are variables that easily confound studies and present a troublesome challenge for the generalization of these studies to other datasets or clinical usage \cite{zhao2019confounder,brookhart2010confounding}.

\begin{figure}
  \centering
  \includegraphics[width=0.48\textwidth,trim=0 0 0 0,clip]{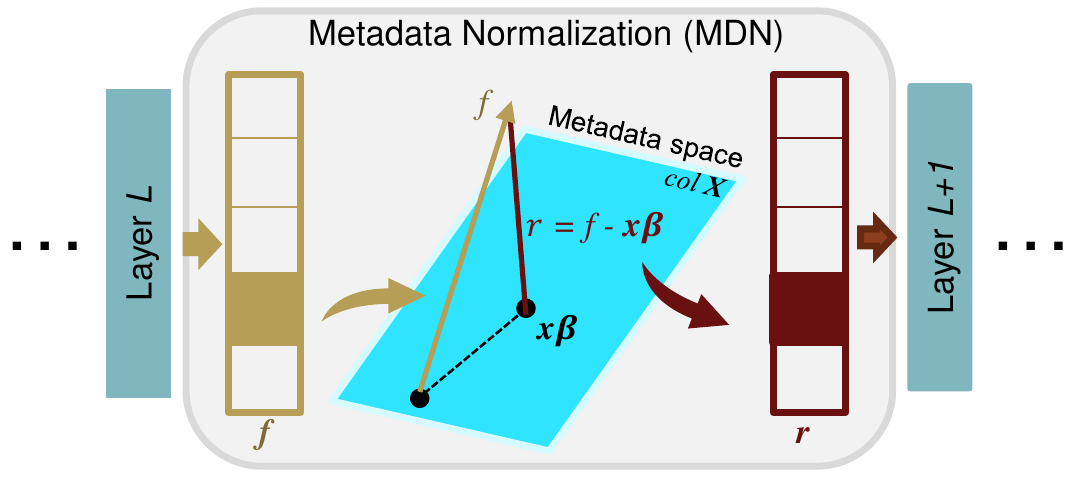}
  \caption{The proposed Metadata Normalization operation. MDN layer takes the learned features from the previous layer ($\boldsymbol{f}$), analyzes the effects of the metadata on them, residualizes such effects, and outputs the distribution corrected features ($\boldsymbol{r}$).}
  \label{fig:mdn_fig1}
  \vspace{-5pt}
\end{figure}

This additional information about training samples is often freely available in datasets (\eg, in medical datasets as patient data) or can be extracted using off-the-shelf models (such as in \cite{buolamwini2018gender,choi2019why}). We refer to them as \textit{metadata}, namely ``data that provides information about other data" \cite{metadata2020}, an umbrella term for additional variables that provide information about training data but are not directly used as model input. The extraneous dependencies between the training data and metadata directly affect the distributions of the learned features; however, typical normalization operations such as BN and GN operate agnostic to this extra information. Instead, current strategies to remove metadata effects include invariant feature learning \cite{xie2017controllable,moyer2018,akuzawa2019adversarial} or domain adaptation \cite{tzeng2017adversarial,ganin2016domain}. 

Traditional handcrafted and feature-based statistical methods often use intuitive approaches based on multivariate modeling to remove the effects of such metadata (referred to as study confounders in this setting). One such regression analysis method \cite{mcnamee2005regression} builds a Generalized Linear Model (GLM) between the features and the metadata (see Fig.~\ref{fig:mdn_fig1}) to measure how much the feature variances are explained by the metadata versus the actual output (\ie, ground-truth label) \cite{zhao2019confounder,adeli2018chained}. The effects of the metadata can then be removed from the features by a technique referred to as ``regressing out" the effects of the extraneous variables \cite{adeli2018chained,brookhart2010confounding,pourhoseingholi2012control}. The application of this GLM-based method to deep 
end-to-end architectures has not yet been explored because it requires precomputed features to build the GLM and is traditionally performed on the dataset prior to training. Thus, this method is inapplicable to vision problems with pixel-level input and local spatial dependencies, which a GLM is unable to model. The key insight we use is that the later layers of a network represent high-level features with which we can build the GLM. In this paper, we extend this widely-explored and seminal regression analysis method by proposing a corresponding operation for deep learning architectures \textit{within a network} to remove the metadata effects from the intermediate features of a network. 

As illustrated in Fig.~\ref{fig:mdn_fig1}, we define a Metadata Normalization (MDN) layer which applies the aforementioned regression analysis as a normalization technique to remove the metadata effects from the features in a network\footnote{Although metadata normalization was previously used to refer to the adjustment of metadata elements into standard formats \cite{koh2018plaster}, we redefine the term as an operation in deep learning architectures.}. Our MDN operation projects each learned feature $\boldsymbol{f}$ of the $L^{th}$ layer to the subspace spanned by the metadata variables, denoted by $\boldsymbol{X}$, by creating a GLM \cite{Neter1996glm,mcnamee2005regression} $\boldsymbol{f}=\boldsymbol{X}\beta+\boldsymbol{r}$, where $\beta$ is a learnable set of linear parameters, $\boldsymbol{X}\beta$ corresponds to the component in $\boldsymbol{f}$ explained by the metadata, and $\boldsymbol{r}$ is the residual component irrelevant to the metadata. The MDN layer removes the metadata-related components from the feature (Fig.~\ref{fig:mdn_fig1}) and regards the residual $\boldsymbol{r}$ as the normalized feature impartial to metadata. We implement this operation in a (mini)batch iterative training setting.

As opposed to BN and its variants that aim at \textit{standardizing the distribution} of the features throughout the training process, MDN focuses on \textit{correcting the distribution} with respect to the chosen metadata variables. When employed in end-to-end settings, this enables deep learning architectures to remove the effects of confounders, protected variables, or biases during the training process. Moreover, the metadata will only correct the distributions if there are such distributions explained by the metadata. On the other hand, if the learned features are orthogonal to the metadata subspace (\ie, features are not biased by the metadata variables), the $\beta$ coefficients will be close to zero and hence will not alter the learning paradigm of the network.

In summary, our work makes the following primary contributions: (1) We propose the Metadata Normalization technique to correct the distribution of the data with respect to the additional, metadata, information; (2) We present a theoretical basis for removal of extraneous variable effects based on GLM and introduce a novel strategy to implement this operator in (mini)batch-level settings; (3) For the cases when output prediction variables are intrinsically correlated with the metadata, we outline a simple extension to MDN to ensure that only extraneous effects of the metadata are removed and not those that pertain to the actual output variables. Our implementation as a simple PyTorch layer module is available at {\small \url{https://github.com/mlu355/MetadataNorm}}. We show the effectiveness of MDN in four different experimental settings, including one synthetic, one image dataset for gender classification from face images, one video scene-invariant action recognition, and one multi-site medical image classification scenario.

\section{Related Work}

\noindent\textbf{Normalization in Deep Learning:} Prior normalization techniques for neural models include Batch Normalization (BN) \cite{ioffe2015batch}, Group Normalization (GN) \cite{wu2018group} and Layer Normalization \cite{ba2016layer} as canonical examples. BN is a ubiquitous mechanism which uses batch statistics to greatly speed up training and boost model convergence. GN and LN apply similar standardization techniques with group and feature statistics to enable smaller batch sizes and improved performance in many settings, especially on recurrent networks. Analogously to these normalization methods, MDN is implemented in batch-level settings on end-to-end networks and uses batch and feature statistics, but differs in how it shifts the distributions of the features (w.r.t. the metadata).

\noindent\textbf{Statistical Methods for Regressing Out Confounders:} Traditional feature-based statistical methods for removing confounders include stratification \cite{diamantidis2000unsupervised}, techniques using Analysis of Variance (ANOVA) \cite{pourhoseingholi2012control}, and the use of multivariate modeling such as regression analysis \cite{mcnamee2005regression} with the statistical GLM method outlined above\cite{zhao2019confounder, adeli2018chained}. However, due to their ineffectiveness in dealing with pixel-level data and dependency on handcrafted features, vision-based tasks and end-to-end methods typically use other techniques to alleviate bias or the effects of study confounders. Common techniques include the use of data preprocessing techniques to remove dataset biases such as sampling bias \cite{zadrozny2004learning} and label bias \cite{jiang2020identifying}. Other methods to remove bias from machine learning classifiers include the use of post-processing steps to enforce fairness on already trained, unfair classification models \cite{feldman2015computational, hardt2016equality,zhao2019confounder}. However, algorithms which decouple training from the fairness enforcement may lead to a suboptimal fairness and accuracy trade-off \cite{woodworth2017learning}. Herein, we apply the ideas behind feature-based confounder removal through regression analysis as a batch-level module, which can be added synchronously to the training process. 

\noindent\textbf{Bias in Machine Learning:}
Bias in machine learning models is an increasingly scrutinized topic at the forefront of machine learning research. The prevalence of bias in large public datasets has been a cause for alarm due to their propagation or even amplification of bias in the models which use them. Recent examples of dataset bias in public image datasets such as ImageNet \cite{yang2020towards}, IARPA Janus Benchmark A (IJB-A) face \cite{klare2015pushing} and Adience \cite{eidinger2014age} have shown that they are imbalanced with mainly light-skinned subjects and that models trained on them retain this bias in their predictions \cite{buolamwini2018gender, shankar2017no}. Bias is prevalent in a wide range of disciplines, such as gender bias in natural language processing via word embeddings, representations, and algorithms \cite{costa2019analysis,sun2019mitigating, bolukbasi2016man} and medical domains \cite{kaushal2020geographic} such as genomics \cite{chadwick2009gender} and Magnetic Resonance Imaging (MRI) analyses \cite{adeli2020deep,eliot2019neurosexism}, in which it is common for data to be skewed toward certain populations \cite{fry2017comparison, stevenson2013sources}. Models trained on such biased settings produce biased predictions or can amplify existing biases. Many approaches have been developed to remove these adverse effects for both qualitative causes (\eg, for fairness) and for quantitative causes (\eg, improving the performance of a model by reducing its dependence on confounding effects). 

Fair representation learning is an increasingly popular approach to learn debiased intermediate representations \cite{zemel2013learning} that has been explored in numerous recent works \cite{hardt2016equality, adeli2019representation, alvi2018turning, louizos2015variational, madras2018learning, wang2019balanced}, with \cite{johndrow2019algorithm, tan2020learning} introducing methods to apply fair and invariant representation learning to continuous protected variables. Recently, adversarial learning has also become a popular area for exploration to mitigate bias in machine learning models \cite{zhang2018mitigating, adeli2019representation}. Our MDN paradigm can interpret this problem as correcting the feature distributions by treating the bias and protected variables as the study metadata, and hence has interesting applications for fair representation learning. In contrast with the domain adaptation and invariant feature learning frameworks, MDN is a layer which easily plugs into an end-to-end learning scheme and is also applicable to continuous protected variables.

\section{Method}

With a dataset including $N$ training samples with prediction labels $(\boldsymbol{I}_i, \boldsymbol{y}_i)$ for $i \in \{1,\ldots,N\}$, we train a neural network with trainable parameters $\boldsymbol{\Theta}$ using  a 2D or 3D backbone, depending on the application (2D for images, and 3D for videos or MRIs). 
MDN layer can be inserted between all convolutional and fully connected layers to correct the distribution of the learned features within the stochastic gradient decent (SGD) framework.

\subsection{Metadata Normalization (MDN) via GLM}
Let $x_i \in \mathbb{R}^K$ be a column vector storing the $K$-dimensional metadata of the $i^{th}$ sample and $\boldsymbol{X}=[\boldsymbol{x}_1,\ldots,\boldsymbol{x}_N]^\top$ be the metadata matrix of all $N$ training samples. Let $\boldsymbol{f}=[f_1,...,f_N]\in\mathbb{R}^N$ be a feature extracted at a certain layer of the network for the $N$ samples. A general linear model associates the two variables by $\boldsymbol{f}=\boldsymbol{X}\beta+\boldsymbol{r}$, where $\beta$ is an unknown set of linear parameters, $\boldsymbol{X}\beta$ corresponds to the component in $\boldsymbol{f}$ explained by the metadata, and $\boldsymbol{r}$ is the residual component irrelevant to the metadata. Therefore, the goal of the MDN layer is to remove the metadata-related components from the feature:
\begin{equation}
    \boldsymbol{r}=\text{MDN}(\boldsymbol{f}; \boldsymbol{X}).
\end{equation}

In this work, we use an ordinary least square estimator to solve the GLM so that the MDN layer can be reduced to a linear operator. Specifically, the optimal $\beta$ is given by the closed-form solution
\begin{equation}
\beta=(\boldsymbol{X}^\top\boldsymbol{X})^{-1}\boldsymbol{X}^\top\boldsymbol{f},
\label{eq:beta}
\end{equation}
and the MDN layer can be written as
\begin{align}
    \boldsymbol{r}&=\boldsymbol{f}-\boldsymbol{X}\beta=\boldsymbol{f}-\boldsymbol{X}(\boldsymbol{X}^\top\boldsymbol{X})^{-1}\boldsymbol{X}^\top\boldsymbol{f}\\
    &=(\textbf{I}-\boldsymbol{X}(\boldsymbol{X}^\top\boldsymbol{X})^{-1}\boldsymbol{X}^\top)\boldsymbol{f}=(\textbf{I}-\textbf{P})\boldsymbol{f}=\textbf{R}\boldsymbol{f}.
    \label{eq:residual_calculation}
\end{align}
Geometrically, $\textbf{P}$ is the projection matrix onto the linear subspace spanned by the metadata (column vectors of $\boldsymbol{X}$) \cite{Basilevsky2005}. The residualization matrix $\textbf{R} \in \mathbb{R}^{N \times N}$ is the residual component orthogonal to the metadata subspace (Fig.~\ref{fig:mdn_fig1}). 

\subsection{Batch Learning}
In a conventional GLM, both $\textbf{X}$ and $\textbf{R}$ are constant matrices defined with respect to all $N$ training samples. This definition poses two challenges for batch stochastic gradient descent. To show this, let $\hat{\textbf{X}} \in \mathbb{R}^{M \times K}$ and $\hat{\boldsymbol{f}} \in \mathbb{R}^{M}$ be the metadata matrix and feature associated with $M$ training samples in a batch. In each iteration, we need to re-estimate the corresponding residualization matrix $\hat{\textbf{R}}\in \mathbb{R}^{M \times M}$, which by Eq.~\eqref{eq:residual_calculation} would require re-computing the matrix inverse $(\hat{\boldsymbol{X}}^\top\hat{\boldsymbol{X}})^{-1}$, a time-consuming task. Moreover, the GLM analysis generally results in sub-optimal estimation of $\beta$ when few training samples are available (\ie, $M\ll N$). To resolve these issues, we further explore the closed-form solution of Eq. \eqref{eq:beta}, which can be re-written as 
\begin{equation}
    \beta = (\boldsymbol{X}^\top\boldsymbol{X})^{-1}\sum_{i=1}^N \boldsymbol{x}_i f_i
    \approx N\boldsymbol{\Sigma}^{-1} \mathbb{E}[\boldsymbol{x}f],
    \label{eq:batch_learning}
\end{equation}
where $\boldsymbol{\Sigma}=\boldsymbol{X}^\top\boldsymbol{X}$ is a property solely of the metadata space independent of the learned features $f$ and $\mathbb{E}[\boldsymbol{x}f]\approx\frac{1}{N}\sum_{i=1}^N \boldsymbol{x}_if_i$. We propose to pre-compute $\boldsymbol{\Sigma}^{-1}$ on all $N$ training samples to derive the most accurate characterization of the metadata space before training. During each training step, we compute the batch-level estimate of the expectation $\mathbb{E}[\boldsymbol{x}f]=\frac{1}{M}\boldsymbol{\hat{X}}^\top\boldsymbol{\hat{f}}$. Hence, the batch estimation of the residualization matrix is
\begin{align}
    \hat{\textbf{R}}&\approx \boldsymbol{\hat{f}} - \boldsymbol{\hat{X}}(N\boldsymbol{\Sigma}^{-1}\mathbb{E}[\boldsymbol{x}f])\\
    &\approx (\boldsymbol{I} - \frac{N}{M}\boldsymbol{\hat{X}}\boldsymbol{\Sigma}^{-1}\boldsymbol{\hat{X}})^\top\boldsymbol{\hat{f}}.
    \label{eq:residualization}
\end{align}
We have rederived our residual solution from Eq.~\eqref{eq:residual_calculation} with the addition of the scaling constant $\frac{N}{M}$ and $(\boldsymbol{\hat{X}^\top\hat{X}})^{-1}$ replaced by our precomputed $\Sigma^{-1}$.

\subsection{Evaluation}
During training, we store aggregated batch-level statistics to use during evaluation, when we may not have a large enough batch to form a reliable GLM solution. Observe from Eq.~\eqref{eq:batch_learning} that $\beta$ is re-estimated in each training batch because the features $\boldsymbol{f}$ are updated after the batch. Since the testing process does not update the model, the underlying association between features and metadata is fixed, so the $\beta$ estimated from the training stage can be used to perform the metadata residualization. To ensure that the $\beta$ estimation can accurately encode the GLM coefficients associated with the entire training set and to avoid oscillation from random sampling of batches), we update $\beta$ at each iteration using a momentum model \cite{sutskever2013importance}:
\begin{equation}
    \beta^k = \eta \beta^k + (1-\eta) \beta^{k-1},
\end{equation}
where $k\in\{1, \dots, \tau\}$ is the batch index and $\eta$ is the momentum constant. During testing, we no longer solve for $\beta$ and instead use the estimate $\beta_\tau$ from the last training batch 
\begin{equation}
    \text{MDN}(\boldsymbol{f};\boldsymbol{X}) = \boldsymbol{f} - \boldsymbol{X}\beta_\tau.
\end{equation} The batch-level GLM solution will approach the optimal group-level solution with increasing batch size, as larger batches produce a better estimate for the dataset-level GLM solution during both training and evaluation.

\subsection{Collinearity between Metadata and Labels}
In more complicated scenarios where confounding effects occur, the metadata not only affects the training input but also correlates with the prediction label. In this case, we need to remove the direct association between $\boldsymbol{f}$ and $\boldsymbol{X}$ while preserving the indirect association created via $\boldsymbol{y}$  \cite{zhaoadeli2020cf-net}. We control for the effect of $\boldsymbol{y}$ by reformulating the GLM as 
\begin{equation}
    \boldsymbol{f}=\boldsymbol{X}\beta_X+\boldsymbol{y}\beta_y+\boldsymbol{r}=\Tilde{\boldsymbol{X}}\Tilde{\beta}+\boldsymbol{r},
\end{equation}
where $\boldsymbol{y}$ is prediction labels vector of the training samples, $\Tilde{\boldsymbol{X}}$ is the horizontal concatenation of $[\boldsymbol{X},\boldsymbol{y}]$, and $\Tilde{\beta}$ is the vertical concatenation of $[\beta_X;\beta_y]$. This multiple regression formulation allows us to separately model the variance within the features explained by the metadata and by the labels, so that we only remove the metadata-related variance from the features. 
To perform MDN in this scenario, we first estimate the composite $\Tilde{\beta}$ in a similar way as in Eq. \eqref{eq:batch_learning} for each batch during training
\begin{equation}
    \Tilde{\beta}
    \approx N\Tilde{\boldsymbol{\Sigma}}^{-1} \mathbb{E}\left[[\boldsymbol{x}f; \boldsymbol{y}f]\right],
\end{equation}
where $\Tilde{\boldsymbol{\Sigma}}$ is the covariance matrix of $\Tilde{\boldsymbol{X}}$ estimated on the whole training population, and the expectation $\mathbb{E}$ is computed on the batch level. Next, unlike the previous MDN implementation in Eq.~\eqref{eq:residualization}, the residualization is now only performed with respect to $\beta_X$

\begin{equation}
\text{MDN}(\boldsymbol{f}; \boldsymbol{X})=\boldsymbol{f}-\boldsymbol{X}\beta_X.
\end{equation} 
Controlling for the labels when fitting features to the metadata preserves the components informative to prediction in our residual and thus in the ensuing features of the network.

\section{Experiments}

We test our method on a variety of datasets covering a diverse array of settings, including both categorical and continuous metadata variables for binary, multi-class, and multi-label classification. For all experiments, our baseline is a vanilla convolutional neural network (2D or 3D CNN), to which we add the proposed MDN to assess its influence on the model learning process. We show that adding MDN to a model can result in improved or comparable prediction accuracy while reducing model dependence on the metadata. The collinearity of the metadata with the labels is handled by adopting the MDN implementation in Section 3.3. We test our method by (1) adding MDN to solely the final \textbf{f}ully-\textbf{c}onnected layers of the network (which we refer to as MDN-FC) and (2) adding MDN to the convolutional layers in addition to the final linear layers (MDN-Conv). For comparison, we add other normalization layers such as Batch Normalization (BN) and Group Normalization (GN) to all convolutional layers of the baseline. 

Computational efficiency is one of the strengths of our method, as there are no learnable parameters due to the closed form solution for batch learning in Section 3.2. Therefore, memory cost is negligible and training time comparable to models without MDN. We used NVIDIA GTX 1080 Ti with 11GB VRAM for image experiments and TITAN RTX with 24GB VRAM for video experiments. 

\subsection{Metrics}
For each of the experiments, we use the squared distance correlation (dcor$^2$) \cite{szekely2007measuring} between our model features and the metadata variables as the primary quantitative metric for measuring the magnitude of the metadata effect on the features. Unlike univariate linear correlation, dcor$^2$ measures non-linear dependency between high-dimensional variables. The lower the dcor$^2$, the less the learned features (and hence the model) are affected by the extraneous variables, with dcor$^2=0$ indicating statistical independence.  The goal is to minimize dcor$^2$ and reduce the dependence between the features and metadata variables. We additionally compute balanced accuracy (bAcc) for all experiments to measure prediction performance. With regard to individual experiments, we compute the following additional metrics: (1) for the GS-PPB experiment, we compute accuracy per shade; (2) for the HVU experiment, we compute mean average precision (mAP) for the action classification task.
\subsection{Synthetic Experiments}

\begin{figure}[!t]
  \centering
    \includegraphics[width=0.7\linewidth]{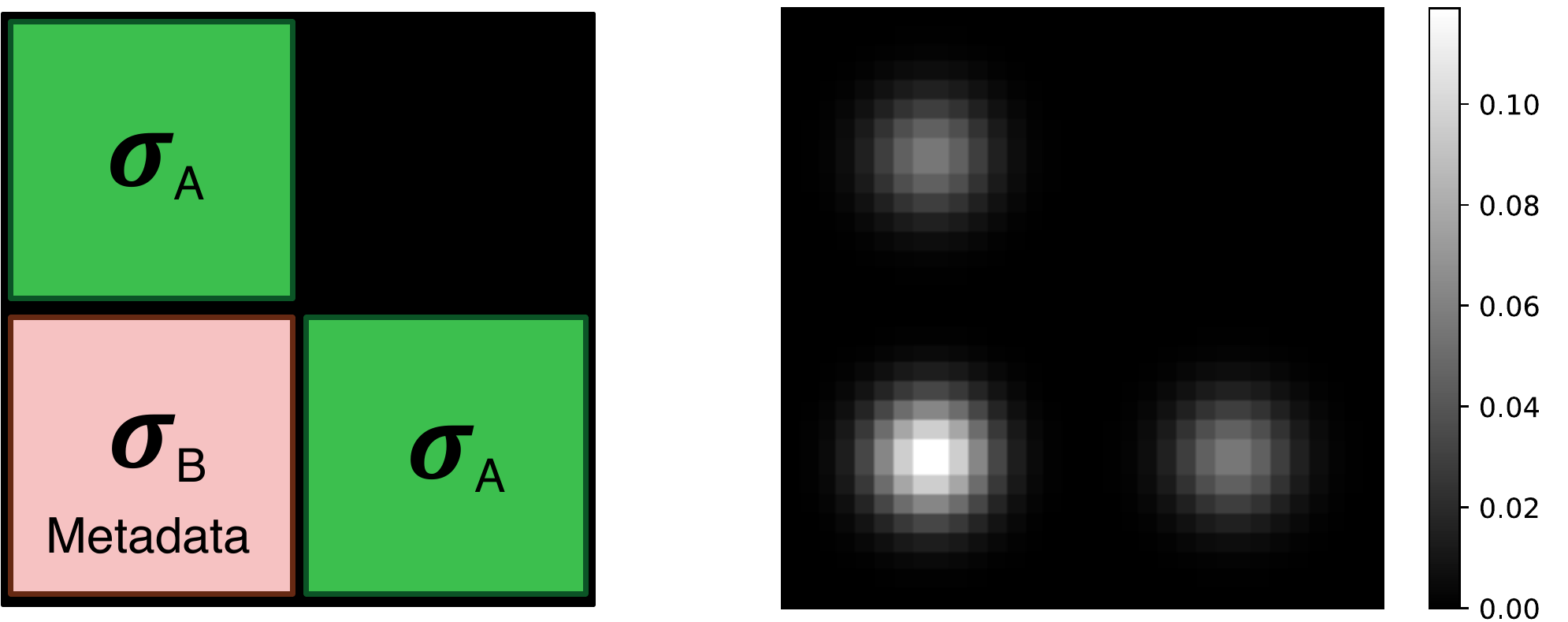}
   \caption{Synthetic data generation process with a generated training sample. The main diagonal Gaussian $\sigma_A$ differentiates the two groups while the off-diagonal Gaussian $\sigma_B$ serves as the metadata.}
   \label{fig:group1}
   \vspace{-5pt}
\end{figure}

\begin{figure*}[!t]
  \centering
  \includegraphics[width=\textwidth]{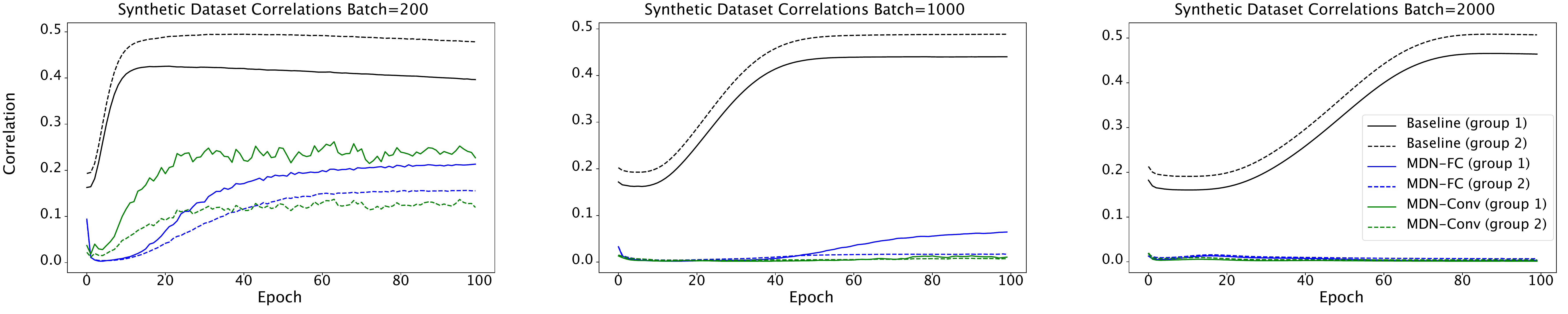}\vspace{-8pt}
  \caption{Effect of MDN across different batch sizes on the synthetic dataset measured in dcor$^2$ (averaged over 100 runs).}
  \label{fig:syn_dcor_comparison}
  \vspace{-5pt}
\end{figure*}

\begin{figure}[t]
  \centering
    \includegraphics[width=0.475\textwidth]{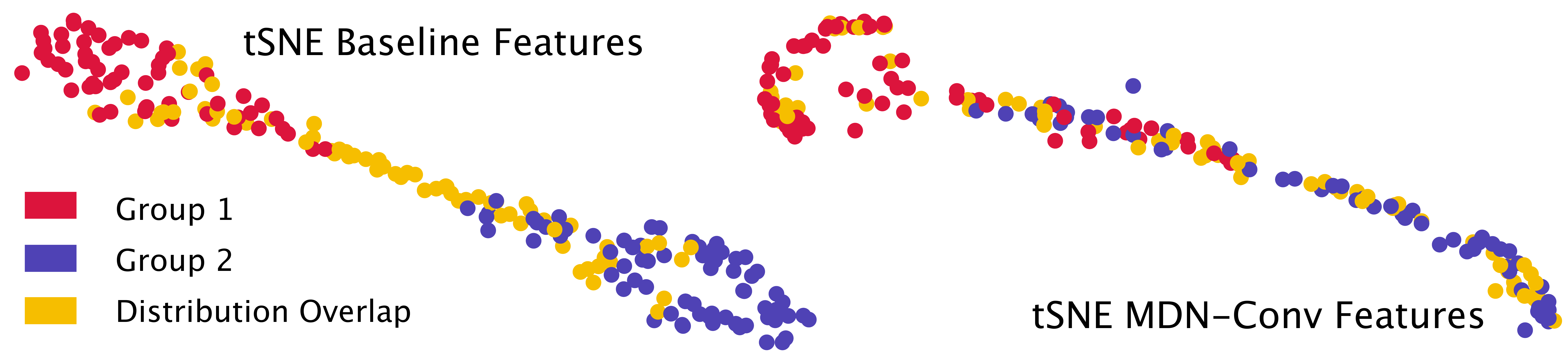}
    \vspace{-5pt}
  \caption{tSNE of features extracted from baseline and MDN-Conv on the synthetic dataset. Samples from the overlapping region $\bm{\mathcal{U}}(3, 4)$ are separable in baseline but not in MDN-Conv.}
  \label{fig:tsne_synthetic}
  \vspace{-5pt}
\end{figure}

The synthetic experiments are constructed as a binary classification task on a dataset of random synthetically generated images comprised of two groups of data, each containing 1000 images of resolution $32 \times 32$ pixels. Each image is generated by 4 Gaussians, the magnitude of which is controlled by parameters $\sigma_A$ (controlling quadrants II and IV) and $\sigma_B$ (controlling quadrant III). Images from Group 1 are generated by sampling $\sigma_A$ and $\sigma_B$ from a uniform distribution $\bm{\mathcal{U}}(1, 4)$, while images from Group 2 are generated with stronger intensities from $\bm{\mathcal{U}}(3, 6)$ (see Figure~\ref{fig:group1}). The difference in $\sigma_A$ between the two groups is associated with the true discriminative cues that should be learned by a classifier, whereas $\sigma_B$ is a metadata variable. Therefore, an unbiased model which is agnostic to $\sigma_B$ should predict the group label purely based on the two diagonal Gaussians without depending on the off-diagonal Gaussian. The overlapping sampling range of $\sigma_A$ between the two groups leads to a theoretical maximum accuracy of 83.33\%.

Our baseline is a simple CNN with 2 convolution/ReLU stacks followed by 2 fully-connected layers of dimension (18432, 84, 1) with Sigmoid activation. We observe the effect of adding MDN to various layers of the baseline: the first (MDN-FC) has one MDN layer applied to the first fully-connected layer, and the second (MDN-Conv) additionally applies MDN to the convolutional layers. 

\begin{table}[t]
    \caption{Comparison of models on the Synthetic Dataset over 100 runs with 95\% CIs for dcor$^2$ (lower is better) and bAcc (closer to 83.3\% is better). Note that the theoretical maximum accuracy of an unbiased model is 83.3\%, so significantly higher values indicate that the model is ``cheating" by using the metadata.}
  \label{tab:synthetic_table}
  \centering
  {\small \renewcommand{\arraystretch}{0.9}
  \setlength{\tabcolsep}{7pt}
  \begin{tabular}{lc|cccc}
   \toprule
    \textbf{Model}       & \textbf{$|\text{Batch}|$} & \textbf{dcor$^2$}  &   \textbf{bAcc}            \\
    \hline \hline
     Baseline \cite{hara2017learning} &     200       &  0.399 $\pm$ 0.014  &   94.1 $\pm$ 0.0  \\ 
              &     1000      &  0.464 $\pm$ 0.004  &   94.1 $\pm$ 0.0    \\ 
              &     2000      &  0.479 $\pm$ 0.005  &   94.1 $\pm$ 0.0     \\ 
     \hline
     BN \cite{ioffe2015batch}&     200       &  0.331 $\pm$ 0.003  &   93.2 $\pm$ 0.1    \\ 
              &     1000      &  0.289 $\pm$ 0.004  &   93.5 $\pm$ 0.1     \\ 
              &     2000      &  0.273 $\pm$ 0.004  &   93.6 $\pm$ 0.1     \\ 
     \hline
     GN \cite{wu2018group}&     200 &  0.368 $\pm$ 0.010 &   94.0 $\pm$ 0.1     \\ 
              &     1000      &  0.399 $\pm$ 0.009  &   94.0 $\pm$ 0.1     \\ 
              &     2000      &  0.435 $\pm$ 0.009  &   94.0 $\pm$ 0.1     \\ 
     \hline
      
     MDN-FC &     200     &  0.189 $\pm$ 0.010  &   90.7 $\pm$ 0.1   \\  
                &     1000    &  0.043 $\pm$ 0.008  &   86.7 $\pm$ 0.7   \\   
                &     2000    &  0.028 $\pm$ 0.012  &   \textbf{82.4 $\pm$ 1.2}  \\    
     \hline
     
     MDN-Conv   &     200     &  0.181 $\pm$ 0.019  &   89.5 $\pm$ 0.8   \\ 
                &     1000    &  0.017 $\pm$ 0.007  &   \textbf{82.8 $\pm$ 0.4}   \\  
                &     2000    &  \textbf{0.003 $\pm$ 0.000 }  &   \textbf{83.4 $\pm$ 0.1}   \\    
    \bottomrule
    \end{tabular}}
    \vspace{-5pt}

\end{table}

Table~\ref{tab:synthetic_table} shows the results of 100 runs of each model over batch sizes 200, 1000, and 2000 with 95\% confidence intervals (CIs). Our baseline achieves 94.1\% training accuracy, significantly higher than the theoretical maximum accuracy of 83.3\%, so it must be falsely leveraging the metadata information for prediction. Similarly, BN and GN produce accuracies around 93\% and 94\%, confirming that neither technique corrects for the distribution shift caused by the metadata. On the other hand, MDN-Conv and MDN-FC produce accuracies much closer to the theoretical unbiased optimum. As batch size increases, both MDN-FC and MDN-Conv decrease in accuracy until hitting the theoretical optimum, which suggests they have completely removed their dependence on the metadata without removing components of features which aid in prediction. MDN-FC reaches the max accuracy at batch size 2000 with 82.4\% and MDN-Conv reaches the max accuracy more quickly, with 82.8\% at batch 1000. We measure dcor$^2$ between the features in the first FC layer and the metadata variable for samples from each group separately (Fig.~\ref{fig:syn_dcor_comparison}) and then record the average in Table~\ref{tab:synthetic_table}. We observe that dcor$^2$ for both MDN models is significantly lower than the lowest baseline dcor$^2$ of 0.399 (Table~\ref{tab:synthetic_table}). The dcor$^2$ decreases as batch size increases; \eg, when using a batch size of 2000, the correlation drops to virtually 0, indicating an exact independence between network features and the metadata variable $\sigma_B$. These results corroborate our expectation from section 3.2 that the batch-level GLM solution will approach the optimal group-level solution with large batches. 

MDN-Conv has superior results to MDN-FC in terms of lower dcor$^2$ and accuracy closer to 83.33\%. This suggests that forming a sequence of linear models by applying MDN to successive layers (after each convolution) may gradually remove nonlinear effects between features and metadata variables. Figure~\ref{fig:tsne_synthetic} shows the tSNE visualization of features extracted from the baseline and from MDN-Conv with 3 groups: samples with kernels sampled from the overlapping $\bm{\mathcal{U}}(3, 4)$ region, which should not be separable without using the metadata, and samples separable into Group 1 and Group 2 using only $\sigma_A$ (with kernels in $\bm{\mathcal{U}}(1, 3)$ and $\bm{\mathcal{U}}(4, 6)$). Our tSNE plot shows that the overlapping region is separable in the baseline features but not in MDN-Conv.

\subsection{Gender Prediction Using the GS-PPB Dataset}

\begin{figure}
  \centering
      \includegraphics[trim=32 10 60 10,clip,width=\linewidth]{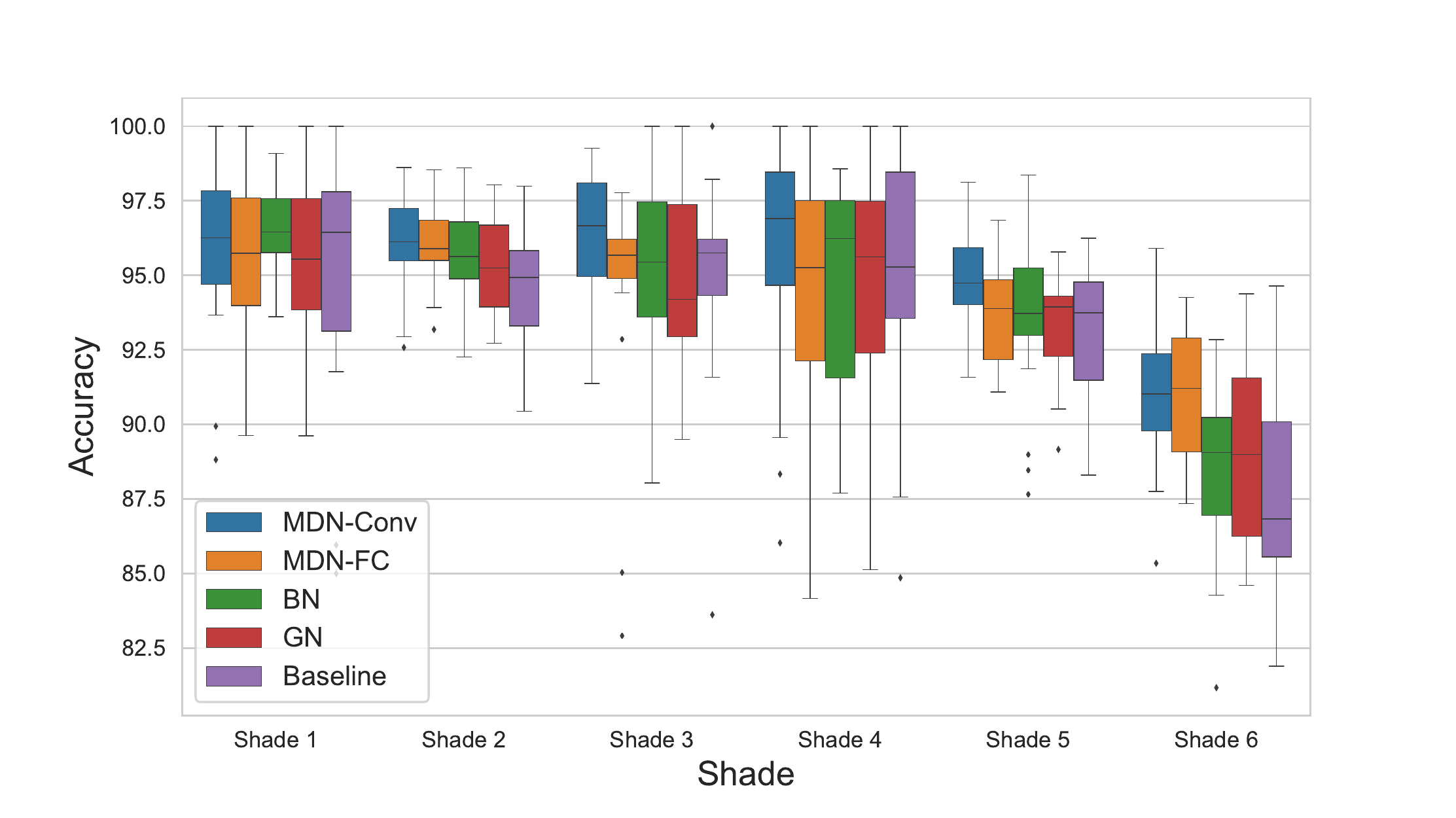}
  \vspace{-15pt}
  \caption{Accuracy by shade on the GS-PPB dataset for different methods pretrained on ImageNet.}
  \label{fig:boxandwhiskers}
  \vspace{-5pt}
\end{figure}

\begin{table}[t]
    \footnotesize
    \caption{Per-shade bAcc and per-class dcor$^2$ on GS-PPB. Results are averaged over 5 runs of 5-fold CV with 95\% Confident Intervals with batch size 250 for all models. Best results are bolded and second best are underlined.}
    \label{ref:per-shade}
    \setlength{\tabcolsep}{2pt}
  \centering
  \begin{tabular}{l|c|c|c|c|c} 
   \toprule
     Shade     & MDN-Conv &                  MDN-FC                      &   Baseline \cite{hara2017learning}            &  BN \cite{ioffe2015batch}   &    GN \cite{wu2018group}\\
    \hline \hline
     1         & \underline{96.2}$\pm$1.1     &   95.9$\pm$1.2               &   95.2$\pm$1.5    &    \textbf{96.6}$\pm$0.6  &     95.6$\pm$1.5  \\
     2         & \textbf{96.2}$\pm$0.6        &   \underline{96.0}$\pm$0.6   &   94.5$\pm$0.8    &    95.7$\pm$0.7           &     95.6$\pm$0.9  \\
     3         & \textbf{96.4}$\pm$0.8        &   94.5$\pm$1.7               &   94.9$\pm$1.2    &    \underline{95.2$\pm$1.1}           &    95.1$\pm$1.4   \\
     4         & \textbf{95.7}$\pm$1.6        &   94.3$\pm$1.8               &   \underline{95.0$\pm$1.6}    &    94.9$\pm$1.4           &   94.6$\pm$2.1    \\
     5         &\textbf{95.1}$\pm$0.7         &   \underline{93.8$\pm$0.8}               &   93.3$\pm$0.8    &    \underline{93.8$\pm$1.0}           &   93.3$\pm$1.0    \\
     6         & \textbf{91.0}$\pm$0.9        &   \underline{90.9}$\pm$1.0   &   87.6$\pm$1.4    &    88.5$\pm$1.0           &    89.0$\pm$1.6   \\
    Avg.& \textbf{95.1}$\pm$0.4    &  \underline{94.2}$\pm$0.5        & 93.4$\pm$0.7     & 94.1$\pm$0.4       &   93.8$\pm$0.5    \\
    \hline
    dcor$^2$ F&  \textbf{0.06}$\pm$0.01& \underline{0.07}$\pm$ 0.01      & 0.14$\pm$0.02  & 0.22$\pm$0.02         &   0.17$\pm$0.01\\
    dcor$^2$ M&  \underline{0.08}$\pm$0.01& {0.09}$\pm$ 0.01      & \textbf{0.05}$\pm$0.01  & 0.10$\pm$0.01         &   0.10$\pm$0.01\\
    Avg.&  \textbf{0.07}$\pm$0.01& \underline{0.08}$\pm$ 0.01      & 0.10$\pm$0.01  & 0.16$\pm$0.01         &   0.13$\pm$0.01\\
    \bottomrule
    \end{tabular}

    \vspace{-5pt}
\end{table}

The next experiment is gender prediction on the face images in the Gender Shades Pilot Parliaments Benchmark (GS-PPB) dataset \cite{buolamwini2018gender}. GS-PPB contains 1,253 face images of 561 female and 692 male subjects, each labeled with a shade on the Fitzpatrick six-point labeling system \cite{fitzpatrick1988validity} from type 1 (lighter) to type 6 (darker). Face detection is used to crop the images to ensure that our classification relies solely on facial features \cite{geitgey2017face}. It has been shown that models pre-trained on large public image datasets such as ImageNet amplify the dependency between shade and target labels due to dataset imbalance \cite{buolamwini2018gender, yang2020towards}. A large discrepancy in classification accuracy of such pre-trained models has been observed between lighter and darker shades, with lowest accuracy in shades 5 and 6. In this experiment, we aim to to reduce the shade bias in a baseline VGG16 backbone model \cite{simonyan2014very} pre-trained on ImageNet \cite{deng2009imagenet} (chosen for its known dataset bias to shade \cite{yang2020towards}) by fine-tuning on the GS-PPB dataset using MDN with shade as the metadata variable. In our VGG16 baseline, we replace the final FC layers with a simple predictor of two FC layers. We test MDN by applying it to the first FC layer (MDN-FC) and additionally to the last convolutional layer (MDN-Conv).

Table~\ref{ref:per-shade} shows prediction results across five runs of 5-fold cross-validation. Per shade accuracies are further visualized in the box plot Figure~\ref{fig:boxandwhiskers}. MDN-Conv and MDN-FC both achieve higher accuracies on the darker shades 5 and 6 with comparable or higher performance for other shades, correcting for the bias in the baseline VGG16 pretrained on ImageNet. Both MDN models also achieve the highest average bAcc and lowest correlations, with MDN-Conv obtaining the highest average bAcc of 95.1\% and the lowest dcor$^2$ on average and for females, with second highest dcor$^2$ for males (F:0.06, M:0.08, Avg:0.07). The baseline achieved slightly lower dcor$^2$ for males but much higher for females and on average (F:0.14, M:0.05, Avg:0.10).

BN and GN increase accuracy when applied to the baseline, which is expected, as they have been shown to improve stability and performance. However, when compared with the MDN models, they produce higher correlation and less robust results. Smaller average accuracy and a much higher dcor$^2$ for females than for males in the non-MDN models indicates that they more heavily leveraging shade for females than for males. This difference is greatly reduced by MDN, which has dcor$^2$ agnostic to gender. Our experiment shows that in settings where features are heavily impacted by metadata, MDN can successfully correct the feature distribution to improve results over normalization methods that only perform standardization using batch or group statistics. 

\subsection{Action Recognition Using the HVU Dataset}
The Holistic Video Understanding (HVU) dataset \cite{hvu} is a large scale dataset that contains 572k video clips of 882 different human actions. In addition to the action labels, the videos are annotated with labels of other categories including 282 scene labels. We use the original split from the paper, with 481k videos in the training set and 31k videos in the validation set. Our task is action recognition with scene as our metadata, aiming to reduce the direct reliance of our model on scene. Action recognition architectures are often biased by the background scene because videos of the same action are captured in similar scenes \cite{Huang2018WhatMA,choi2019why}. These architectures may capture the easier scene cues rather than the harder-to-understand movement cues that define the action in time, which can reduce generalizability to unseen cases.

\begin{figure}
  \centering
      \includegraphics[width=0.95\linewidth]{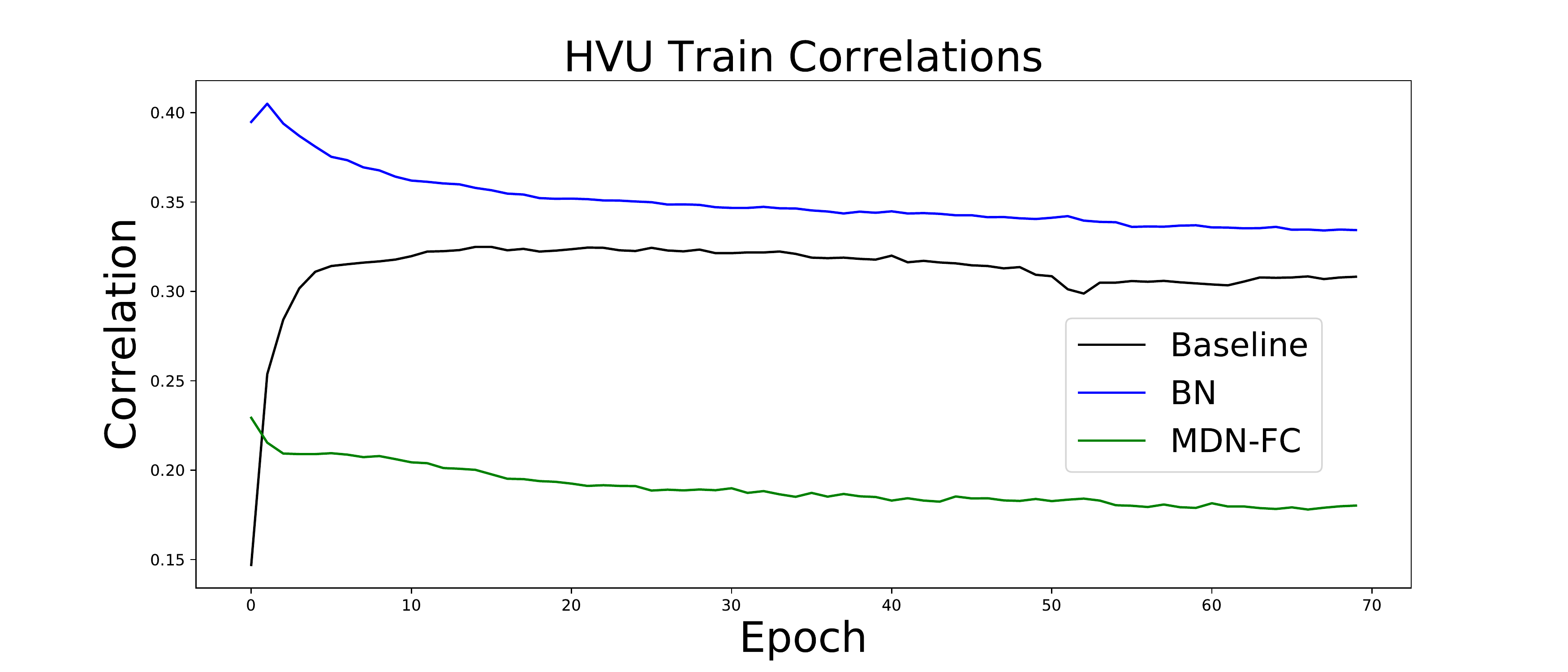}
  \caption{Comparison of train dcor$^2$ for MDN-FC, Baseline and BN on HVU pretrained on the Kinetics-700 Dataset}
  \label{fig:hvu_train_dcor}
  \vspace{-5pt}
\end{figure}

We use a 3D-ResNet-18 \cite{hara2017learning} architecture pretrained on the Kinetics-700 dataset \cite{carreira2017quo} as our BN model and our baseline is the same with no normalization (BN layers removed). MDN is added to the final FC layer of the baseline before the output layer (MDN-FC). The metadata variables are one-hot vector encodings of the top 50 scenes by occurrence. The model is fine-tuned on the HVU dataset until validation accuracy converges, at around 70 epochs. Figure~\ref{fig:hvu_train_dcor} and Table~\ref{ref:HVU} show that while the baseline model increases in dcor$^2$ during training, MDN and BN decrease, even though BN has higher dcor$^2$. MDN displays the lowest dcor$^2$ by far of 0.182, so we have clearly succeeded at reducing the dependence between our model features and scene. However, both MDN and BN experience slightly lower mAP than the baseline without normalization. This is not surprising since scenes may provide information about the action itself, separating cases where the model directly uses scene for prediction and indirectly does so with action as an intermediate dependency. Thus, removing scene dependence may hurt model performance measured by metrics such as mAP by either preventing it from ``cheating" by directly using the scene for prediction, or removing useful components of the features. We have demonstrated that MDN successfully removes direct reliance of our model on scene, but further exploration is needed to fully interpret the effect on model performance and generalizability. 

\begin{table}[!t]
    \caption{Action recognition performance on the HVU dataset.}
  \label{ref:HVU}
  \centering
  \setlength{\tabcolsep}{5pt}
  \begin{tabular}{lcccccc}
  \toprule
    Model                                                 & dcor$^2$         & mAP (\%) \\
    \midrule
     BN (3D-ResNet-18) \cite{ioffe2015batch}               & 0.335           & 39.7    \\ 
     Baseline (3D-ResNet-18 - BN) \cite{hara2017learning}  & 0.307           & 42.4    \\ 
     MDN-FC (3D-ResNet-18 + MDN)                           & 0.182           & 40.3   \\ 
    \bottomrule
    \end{tabular}
    \vspace{-5pt}
\end{table}

This video action recognition experiment is also particularly challenging due to the large dimensionality of video input. This is a common problem in video recognition tasks for batch-level operations such as BN. In our case with batch size 256, larger batch sizes would provide better estimates of the GLM parameters. Several prior works \cite{defossez2017adabatch,chen2018lag, singh2020filter} have proposed work-around solutions by calculating aggregated gradients from several batches or aggregated batch-level statistics to virtually increase batch size. This area requires further study and we anticipate that implementing such strategies may improve results further. 

\subsection{Classification of Multi-Site Medical Data}
\begin{table}[!b]
\caption{Multi-site multi-label disease diagnosis classification based on 3D MRIs and accuracy scores with respect to the target (UCSF) dataset. Recall rate (accuracy) for each cohort, bAcc of all cohorts, standard deviation of per-cohort recall, and distance correlation between the learned features and dataset labels.}
\centering
{ \small
  \setlength{\tabcolsep}{4pt}
\begin{tabular}{l|cc|c|c|c|c}
\toprule
 \multirow{2}{*}{Class} & \multicolumn{2}{c|}{\# of Subjects} & \multirow{2}{*}{Baseline}& \multirow{2}{*}{BN \cite{ioffe2015batch}}   & \multirow{2}{*}{GN \cite{wu2018group}}   & \multirow{2}{*}{MDN} \\ 
 & Total & UCSF & & & \\ \hline 
 CTRL  & 460 & 156          & 8.3\% &  9.6\% &   14.1\% & 41.7\%       \\ 
 HIV   & 112 & 37           & 63.0\% &  62.2\% &  66.9\% &   32.4\%         \\ 
 MCI   & 732 & 335          & 37.8\% & 54.6\% &    51.4\% &  73.7\%      \\ 
 HAND  & 145 & 145          & 42.1\% &  55.9\% &   52.4\% &  42.0\%\\ 
\hline \hline
\multicolumn{3}{r|}{Overall bAcc} & 37.8\% & 45.6\% & 46.1\% & \textbf{47.5\%}  \\
\multicolumn{3}{r|}{Standard Deviation} &22.5\%& 25.2\% & 22.5\% & \textbf{18.1\%}  \\
\multicolumn{3}{r|}{dcor$^2$} & 0.26 & 0.30 & 0.34 & \textbf{0.06} 
\vspace{-5pt}
\end{tabular}}

\label{tab:ucsf_acc}
\end{table}

\begin{figure}[!t]
  \centering
\setlength{\tabcolsep}{1pt}
{\footnotesize
\begin{tabular}{cc}
   \includegraphics[trim=8 5 92 25,clip,width=0.49\linewidth]{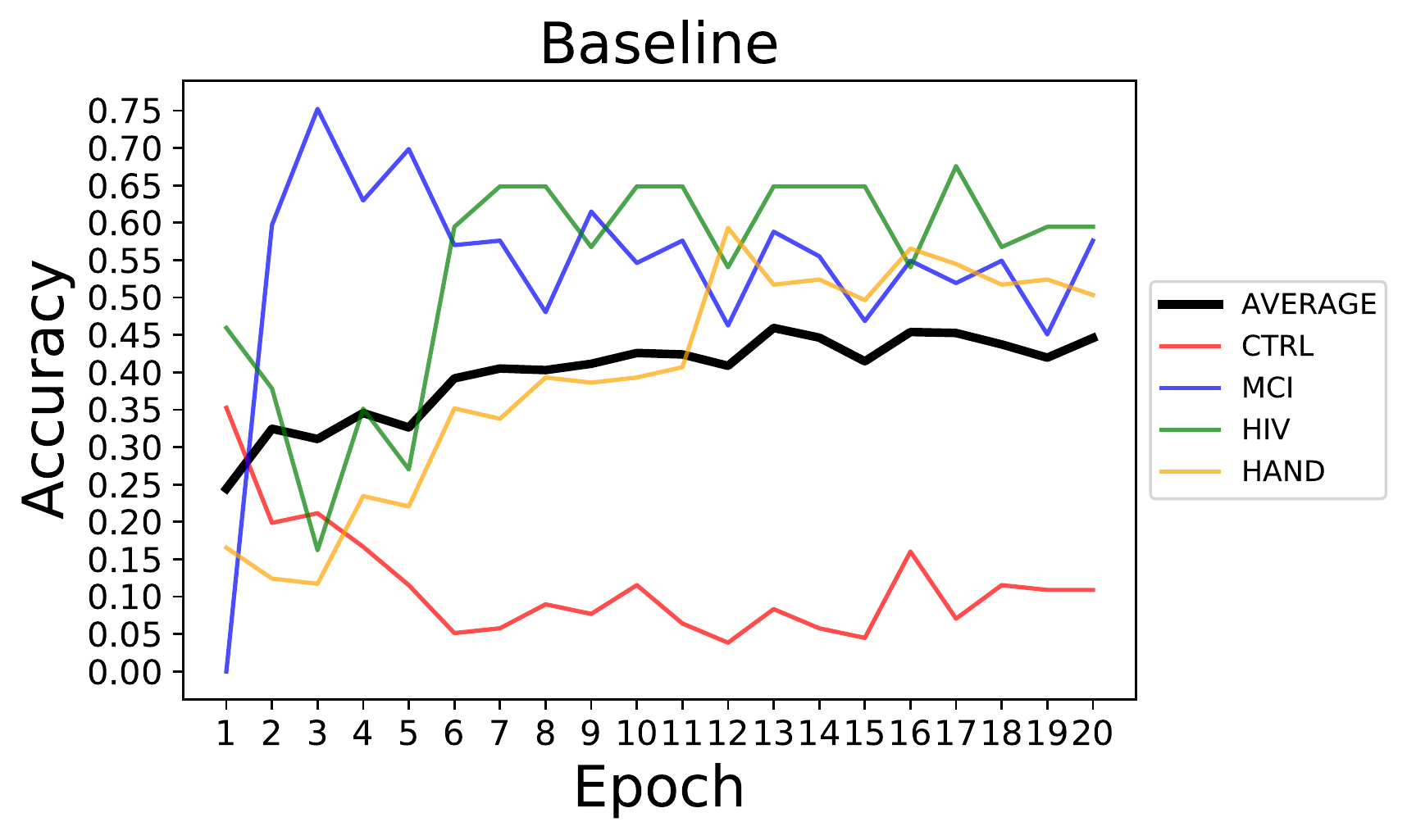}
   & \includegraphics[trim=26 5 5 25,clip,width=0.475\linewidth]{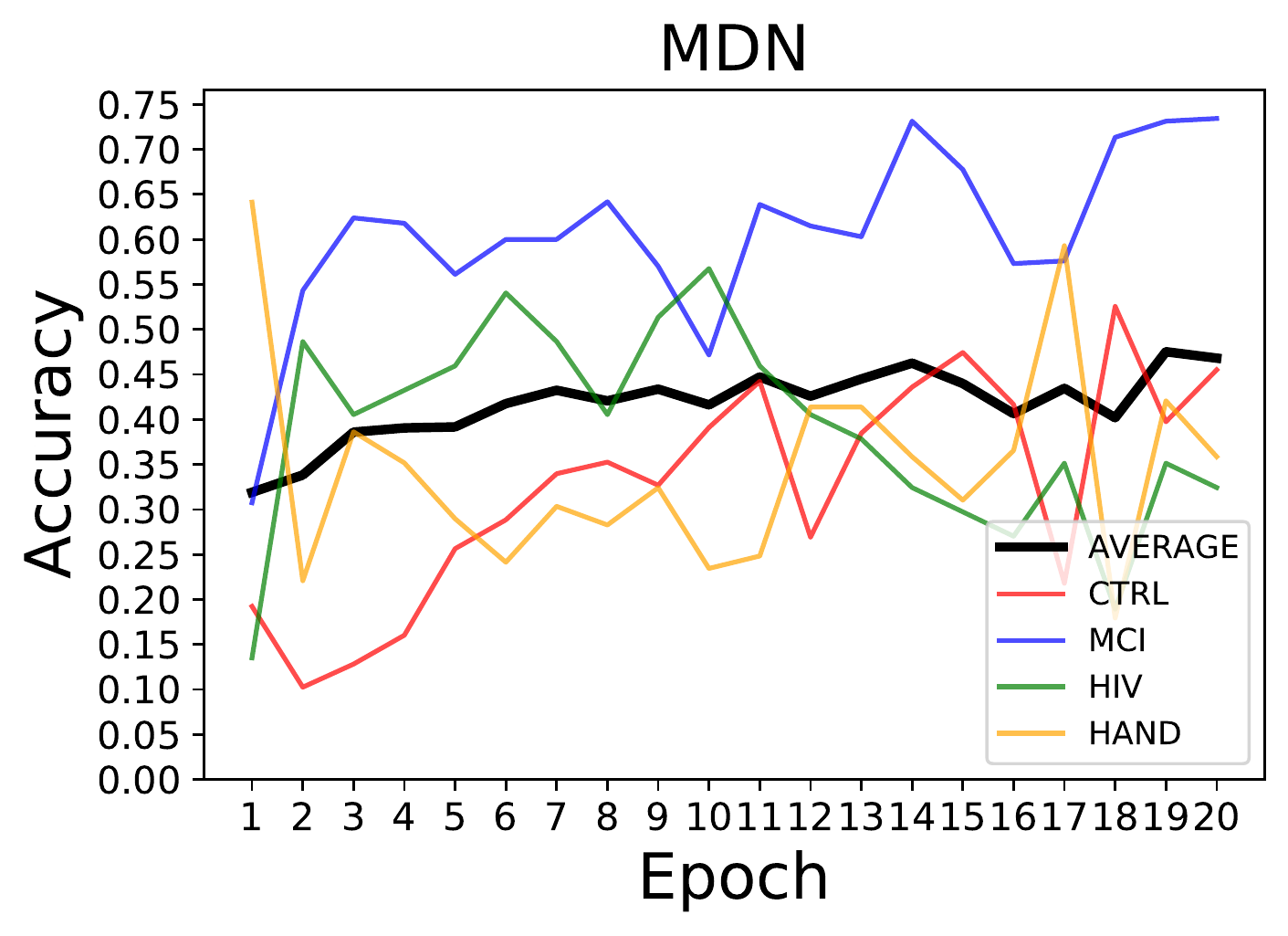} \\
   (a) BN & (b) MDN
\end{tabular}}
    \caption{Prediction accuracy of each cohort and the balanced accuracy over the 4 cohorts on the testing folds (averaged over the 5 folds) versus training iterations.}
  \label{fig:ucsf_acc}
\vspace{-5pt}
\end{figure}

\begin{figure}[t]
  \centering
\begin{picture}(240,120)
\put(0,0){\includegraphics[width=0.475\textwidth,trim=0 0 0 35,clip]{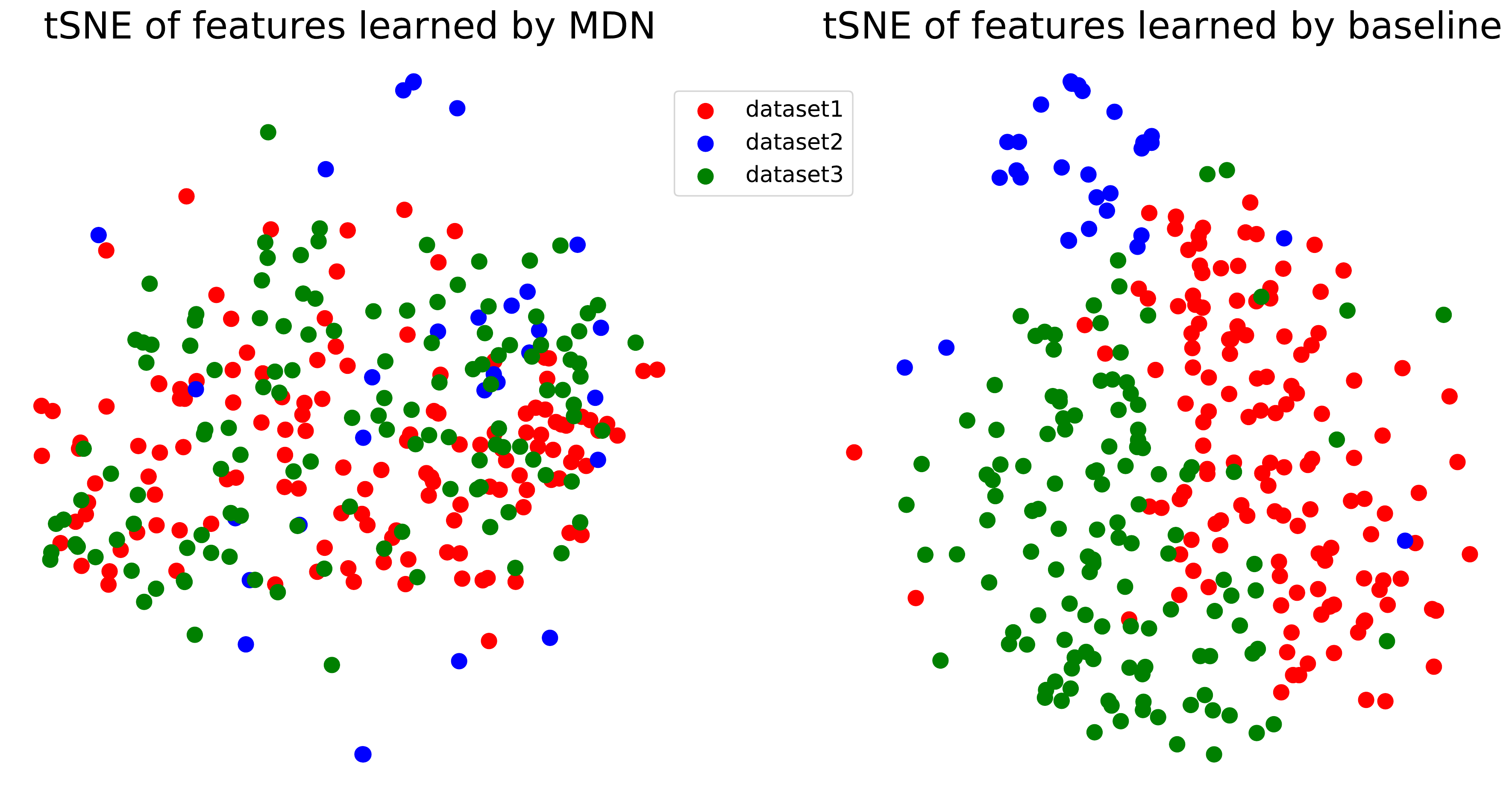}}
\put(10,10){(a) MDN}
\put(120,10){(b) BN}
\end{picture}    \vspace{-15pt}
  \caption{tSNE visualization of the features extracted from the convolutional layers with BN or with MDN. Each data point is color-coded based on the dataset label.}
  \vspace{-5pt}
  \label{fig:ucsf_tsne}
\end{figure}

The last experiment is diagnostic disease classification of 4 cohorts of participants based on their T1-weighted 3D MRI scans. The 4 cohorts are healthy control (CTRL) subjects, subjects that show Mild Cognitive Impairment (MCI), those diagnosed with Human Immunodeficiency Virus (HIV) infection, and subjects with HIV‐Associated Neurocognitive Disorder (HAND). Since HAND is a comorbid condition that combines the characteristics of HIV and MCI, the classification is formulated as a multi-label binary classification problem, where we predict for each subject whether 1) the subject has MCI diagnosis; and 2) the subject is HIV-positive. The HAND patients are positive for both labels and the CTRLs are negative for both.

The T1-weighted MR images used in this study were collected at the Memory and Aging Center, University of California - San Francisco  (UCSF; PI: Dr.~Valcour) \cite{zhang2016extracting} shown in Table \ref{tab:ucsf_acc}. Since the number of subjects is relatively small, especially for the HIV cohort ($N=37$), we augment the training dataset with MRI scans collected by the Neuroscience Program, SRI International (PI: Dr.~Pfefferbaum), consisting of 75 CTRLs and 75 HIV-positive subjects \cite{adeli2018chained}, and by the public Alzheimer's Disease Neuroimaging Initiative (ADNI1) \cite{petersen2010alzheimer}, which contributed an additional 229 CTRLs and 397 MCI subjects. To perform classification on such multi-site data, the source of the data (dataset label) becomes the metadata, which is parameterized here by one-hot encodings. Medical imaging datasets acquired in multiple sites with different scanning protocols is a core challenge for machine learning algorithms in medicine, \cite{ma2018classification,wachinger2020detect}, as different scanning protocols lead to different image formations. Differing class formations across sites (as in this experiment) creates a simple undesirable cue for the model to leverage during prediction as a confounder. We corrected the features distribution by deeming the acquisition site as our metadata variable and employing our MDN operation.

The baseline classification model consists of a feature extractor and a classifier. We designed the feature extractor as 4 stacks of $3\times 3 \times 3$ convolution/ReLU/BN/max-pooling layers with dimension (16, 32, 64, 32). The classifier consists of 2 fully connected layers with dimension (2048, 128, 16). We construct each batch by sampling 10 subjects from each cohort of each dataset (with replacement). The model accuracy is evaluated by 5-fold cross-validation with respect to the 4 cohorts of UCSF, which is the primary goal of the experiment. We train the model for 20 epochs until the bAcc on the testing folds (averaged over the 5 testing folds) converges. 
We then rerun the experiment by replacing the BN layers in the baseline model with MDN layers.

We observe from Table \ref{tab:ucsf_acc} and Fig.~\ref{fig:ucsf_acc} that MDN improves bACC for the multi-label prediction compared to the baselines with BN and GN. The baseline models exhibit highly imbalanced predictive power among the 4 cohorts reflected by the large discrepancy in per-cohort recall (std of 25.2\% and 22.5\%). This is in part because the 3 datasets represent distinct cohort constructions (\eg, ADNI only contains MCI, but no HIV) so the multi-domain feature distribution is likely to bias the discriminative cues related to the neurological disorders. This, however, is not the case for MDN, which successfully reduces the accuracy discrepancy among cohorts (std of 18.1\%). The CTRL group received an especially high increase in bAcc, from 9.6\% in BN and 14.1\% in GN to 41.7\% in MDN. 

The reduced dataset bias is also evident in the distance correlation analysis, which examines the dependency between the features extracted from the convolutional layers and the dataset label. Table \ref{tab:ucsf_acc} records the average dcor$^2$ derived over the 5 testing folds, and MDN achieves a significantly lower metric than the baseline model. Qualitatively, we randomly select a testing fold and use t-SNE to project the features learned by the two models into a 2D space and color-code the data point by their dataset label (Fig.~\ref{fig:ucsf_tsne}). The features are clearly clustered by dataset assignment for the baseline model, whereas this adverse clustering effect is significantly reduced after MDN.

\section{Conclusion}

We presented a novel normalization operation for deep learning architectures, denoted by Metadata Normalization (MDN). This operation, used in end-to-end settings with any architecture, removes undesired extraneous relations between the learned features of a model and the chosen metadata. MDN extends traditional statistical methods to deep learning architectures as a network layer that corrects the distribution shift of model features from metadata, differing from BN and its variants that only standardize the features.  Therefore, it can effectively combat bias in deep learning models as well as remove the effects of study confounders in medical studies. Our results on four diverse datasets have shown that MDN is successful at removing the dependence of the learned features of a model on metadata variables while maintaining or improving performance. 

\vspace{2pt}
\noindent\textbf{Acknowledgements.} This study was partially supported by NIH Grants (AA017347, MH113406, and MH098759), Schmidt Futures Gift, and Stanford Institute for Human-Centered AI (HAI) AWS Cloud Credit.

{\small
\bibliographystyle{ieee_fullname}
\bibliography{refs}
}

\end{document}